\pdfoutput=1

\documentclass[11pt]{article}
\usepackage[dvipsnames]{xcolor}

\usepackage[preprint]{acl}

\usepackage{times}
\usepackage{latexsym}

\usepackage[T1]{fontenc}

\usepackage[utf8]{inputenc}

\usepackage{microtype}

\usepackage{inconsolata}
\usepackage{todonotes}

\usepackage{booktabs}
\usepackage{tabularx,ragged2e}
\usepackage{multirow}
\usepackage{array}

\usepackage{graphicx}
\graphicspath{ {./figures/} }

\usepackage{pgf} 
\usepackage{colortbl}

\newcommand{\gradientcell}[7]{%
    \ifdimcomp{#1pt}{>}{#3 pt}{\cellcolor{#5!100.0!#4!#6}\hspace*{-4.0pt}#7 \hspace*{-4.0pt}}{%
    \ifdimcomp{#1pt}{<}{#2 pt}{\cellcolor{#5!0.0!#4!#6}\hspace*{-4.0pt}#7 \hspace*{-4.0pt}}{%
         \pgfmathparse{int(round(100*(#1/(#3-#2))-(#2 *(100/(#3-#2)))))}%
        \xdef\tempa{\pgfmathresult}%
        \cellcolor{#5!\tempa!#4!#6}\hspace*{-4.0pt}#7\hspace*{-4.0pt}%
    }}
}
\newcommand{\corr}[2]{\gradientcell{#1}{0.0}{0.79}{Cyan}{Yellow}{60}{#2}}
\newcommand{\C}[1]{\pgfmathparse{abs(#1)}\corr{\pgfmathresult}{#1}}

\title{Beyond Literal Token Overlap: Token Alignability for Multilinguality}

\author{
  \textbf{Katharina Hämmerl\textsuperscript{1,2}},
  \textbf{Tomasz Limisiewicz\textsuperscript{3}},
  \\
  \textbf{Jindřich Libovický\textsuperscript{3}},
 \textbf{Alexander Fraser\textsuperscript{4,2}}
\\
  \textsuperscript{1}Centre for Information and Language Processing, LMU Munich
\\
  \textsuperscript{2}Munich Center for Machine Learning
\\
  \textsuperscript{3}Faculty of Mathematics and Physics, Charles University, Czech Republic
\\
  \textsuperscript{4}Technical University of Munich, Germany
\\
\small{
    \textbf{Correspondence:} \texttt{haemmerl [at] cis [dot] lmu [dot] de}
  }
}

\begin{document}
\maketitle
\begin{abstract}

Previous work has considered token overlap, or even similarity of token distributions, as predictors for multilinguality and cross-lingual knowledge transfer in language models.
However, these very literal metrics assign large distances to language pairs with different scripts, which can nevertheless show good cross-linguality.
This limits the explanatory strength of token overlap for knowledge transfer between language pairs that use distinct scripts or follow different orthographic conventions.
In this paper, we propose \textit{subword token alignability} as a new way to understand the impact and quality of multilingual tokenisation.
In particular, this metric predicts multilinguality much better when scripts are disparate and the overlap of literal tokens is low.
We analyse this metric in the context of both encoder and decoder models, look at data size as a potential distractor, and discuss how this insight may be applied to multilingual tokenisation in future work.
We recommend our subword token alignability metric for identifying optimal language pairs for cross-lingual transfer, as well as to guide the construction of better multilingual tokenisers in the future.
We publish our code and reproducibility details\footnote{\url{https://github.com/KathyHaem/token-alignability}}.
\vspace{5pt}
\end{abstract}

\section{Introduction}

\begin{figure}[t!]
    \centering
    \includegraphics[width=\linewidth]{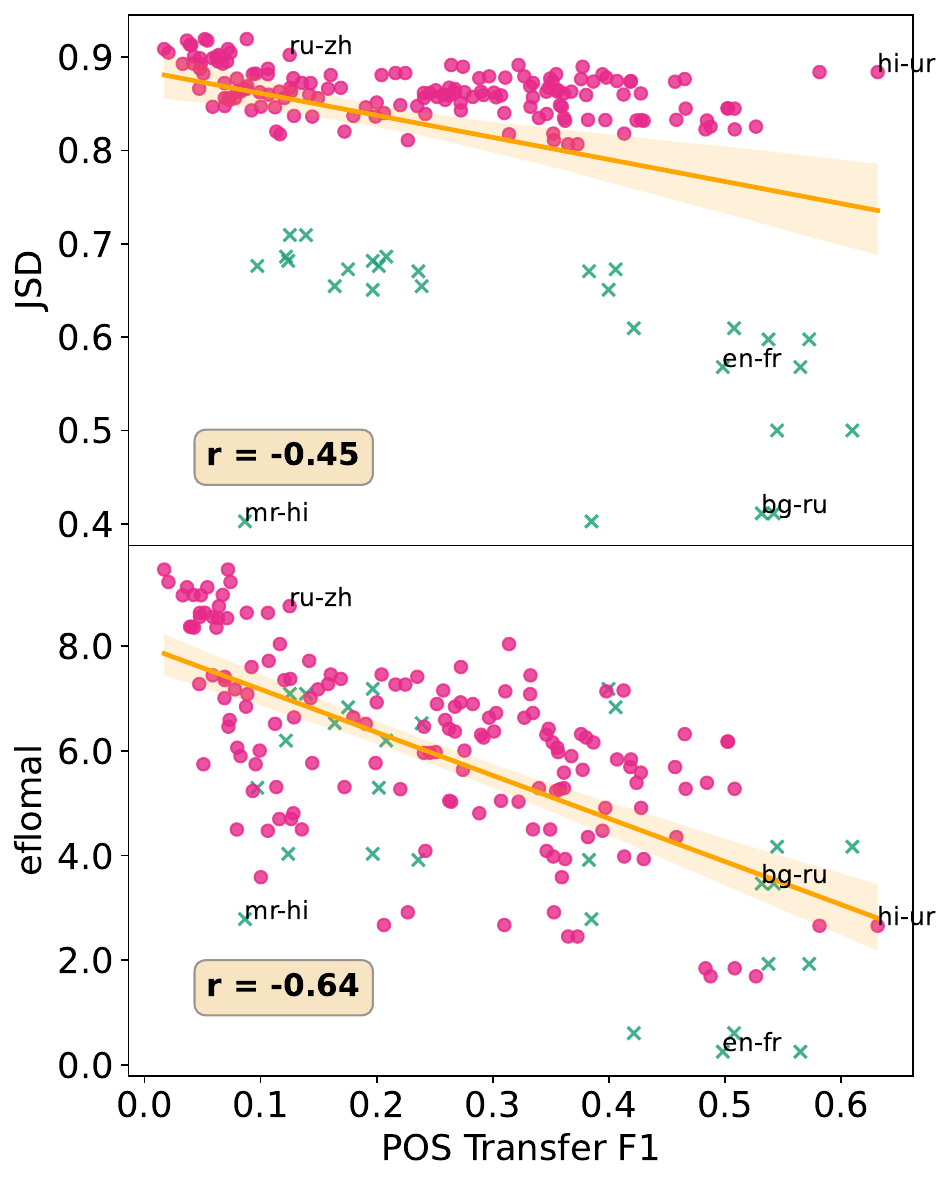}
    \caption{Eflomal score (bottom), a measure of token alignability, predicts downstream transfer performance better than the previous metric of distributional token overlap (top).
    The difference is especially stark for language pairs with \textcolor{magenta}{different scripts ($\bullet$)}, compared to language pairs with the \textcolor{teal}{same script ($\times$)}. The orange line
    shows the linear fit across all included pairs.
    }
    \label{fig:jsd-vs-eflomal}
\end{figure}

Highly multilingual language models have received plenty of research attention in recent years.
\textit{Cross-lingual alignment} of representations, that is, the similar representation of similar meanings regardless of input language
\citep{libovicky-etal-2020-language, hammerl-etal-2024-understanding}, as well as good downstream cross-lingual transfer ability
(\citealp[cf.][]{huang-etal-2019-unicoder, schuster-etal-2019-cross-lingual, hu2020xtreme, pham-etal-2024-unibridge}, etc.), have been considered desirable properties for such models.
Representation alignment is typically seen as a key contributing factor to transfer ability, which in turn enables efficient handling of numerous task-language combinations.
A number of papers have asked when and why information is shared across language boundaries in multilingual models and enables cross-lingual transfer (\citealp{dufter-schutze-2020-identifying,deshpande-etal-2022-bert,limisiewicz-etal-2023-tokenization,hua-etal-2024-mothello, schäfer2024rolelanguageimbalancecrosslingual}, inter alia).

Token overlap, i.e., the occurrence of identical tokens in the corpora of multiple languages, has been shown to affect the cross-lingual capabilities of models \cite{wu-dredze-2019-beto}.
Another approach is to compare the distributions of token literals in parallel corpora (\citealp{limisiewicz-etal-2023-tokenization}).
Still, both metrics have a crucial limitation:
They cannot explain why related languages with different scripts are well-aligned by the models (see~\S~\ref{subsec:related-tokenisation}).

Here, we propose another angle: token alignability.
This concept captures the intuition that models may rely on statistical correspondences between subword tokens (`token alignment') that are more nuanced than literal string matching.
From token alignments produced by a statistical word aligner, we derive two kinds of \textit{token alignability scores} for any language pair in a multilingual tokeniser: one directional, one symmetrised (\S~\ref{subsec:alignability-calc}).

We compute correlations of these scores both to downstream transfer performance on classification and sequence labelling
tasks~(cf.~\S~\ref{subsec:def-transfer}),
and to measures of cross-lingual alignment in the model representations (cf.~\S~\ref{subsec:def-cla}).
Our primary object of study is a set of small encoder models trained with several different multilingual tokenisers (BPE, Unigram, and `TokMix').
Furthermore, we also consider recent larger, pre-trained decoder models.
In addition to showing that token alignability is a better predictor of downstream cross-lingual transfer than distributional overlap (\S~\ref{subsec:main-results}), we consider the impact of pre-training data size (\S~\ref{subsec:data-size}),
and show the correlation of token alignability with representation alignment inside the model
(also~\S~\ref{subsec:main-results}).
Finally, we discuss how this insight may be applied to future multilingual tokenisers (\S~\ref{sec:future}).

\section{Related Work}

Subword tokenisation is currently the standard input processing approach of language models, with BPE \citep{sennrich-etal-2016-neural} and UnigramLM \citep{kudo-2018-subword} being the most common algorithms for deriving these tokens.
However, there has been increased interest in recent years in addressing limitations of the subword token paradigm \citep[e.g.,][]{alkaoud-syed-2020-importance, hofmann-etal-2022-embarrassingly, schmidt2024tokenizationcompression}
or even moving beyond it \citep[e.g.,][]{xue-etal-2022-byt5, mofijul-islam-etal-2022-vocabulary}.

\subsection{Influence of tokenisers on cross-linguality}\label{subsec:related-tokenisation}

Most relevant for our purposes are measurements of tokeniser properties
\citep[e.g.,][]{zouhar-etal-2023-tokenization, batsuren2024evaluatingsubwordtokenizationalien}, particularly for multilingual language models.
\citet{limisiewicz-etal-2023-tokenization} measure the distance of a language pair's token vocabulary via
divergence of the two token distributions.
They find that this kind of `soft overlap' measure correlates well with downstream transfer performance, with an important caveat: the observed correlations are strong for language pairs with the same script, but weaker for pairs with different scripts.
This is because of how the metric is calculated:
The occurrences of subword tokens are counted on each side of a parallel corpus, giving a distribution per language.
Then, Jensen-Shannon-Divergence (\textbf{JSD}; \citealp{lin2006divergence}) is calculated, which gives a symmetrized distance between the two distributions of subword tokens.
The literal matching limits the predictive power of their metric for pairs with different scripts---for instance, Hindi and Urdu are known to be related languages written in different scripts.
Transfer between them works well, while the computed distance is large.

\subsection{Word Alignment in MT}

\textit{Alignment}, in the sense used in statistical Machine Translation (MT) \citep{brown93:tmo} is
a mapping between parallel sentences, showing which tokens are translations of one another and how often they correspond across whole corpora.
The original intuition behind attention is that it finds this kind of mapping in a contextualised manner \citep{bahdanau2015attention},
whereas statistical word aligners
(we use eflomal; \citealp{oestling-tiedemann-2016-efficientWA}) give a discrete mapping.

\section{Methodology}\label{sec:methods}

Our central analysis relies on rank correlations, showing which tokeniser metrics (\S~\ref{subsec:jsd}, \S~\ref{subsec:alignability-calc}) are more predictive of downstream cross-lingual transfer (\S~\ref{subsec:def-transfer}) and cross-lingual alignment of representations (\S~\ref{subsec:def-cla}).
We ensure that within each task, the metrics are always compared over the same set of language pairs.

\subsection{Distributional/Soft Overlap (JSD)}\label{subsec:jsd}

We measure soft overlap between the token distributions of two tokenised corpora.
We follow the setting used by \citet{limisiewicz-etal-2023-tokenization} and outlined in \S~\ref{subsec:related-tokenisation}, but we compute it on the FLORES-200 corpus \cite{guzman-etal-2019-flores,goyal-etal-2022-flores,nllb2022} for comparison with our proposed metrics.
This score is symmetric between both directions of a language pair.
A lower score corresponds to a smaller distance and is thus better.

\subsection{Token alignability of a language pair}\label{subsec:alignability-calc}

We define the \textit{token alignability score} for a language pair based on the symmetrised word alignment of one parallel corpus after training the tool on another.
To train the priors, we use OPUS-100 data \cite{tiedemann-2012-parallel,zhang-etal-2020-improving} 
for en-xx language pairs, and subsets of MultiCCAligned \cite{tiedemann-2012-parallel,el-kishky-etal-2020-ccaligned}
for non-English language pairs.
Seee Appendix~\ref{sec:app-langs} for a breakdown of language pairs.
For each training corpus, we take up to 300k sentence pairs.

As our test corpus, we use FLORES-200 \cite{guzman-etal-2019-flores,goyal-etal-2022-flores,nllb2022} because of its multi-parallel nature and less noise compared to MultiCCAligned.
Following \citet{vazquez-etal-2019-university},
we run a statistical (discrete) word aligner (specifically \textbf{eflomal}; \citealp{oestling-tiedemann-2016-efficientWA}) on the test corpus with a single iteration. 
Based on the final symmetrised alignment over the test corpus, we can determine:
\begin{itemize}
    \item[a)] The \textit{proportion of 1-1 token alignments} (higher is better), i.e., the rate of subword tokens in the source language text with a one-to-one correspondence to subword tokens of the target language text. We take this measure per direction, since it can be markedly lower if the source language is over-segmented.
    \item[b)] The \textit{eflomal score} (lower is better), which represents the tool's estimation of the ``maximum unnormalized log-probability of links
    in the last sampling iteration'' \citep{vazquez-etal-2019-university}, given the learned priors over the subword vocabulary and corpus. We average this score over both directions of a language pair.
\end{itemize}

\subsection{Downstream cross-lingual transfer}\label{subsec:def-transfer}

We were able to obtain model instances with several distinct tokenisers (BPE, Unigram, TokMix), and results for downstream cross-lingual transfer, from the authors of \citet{limisiewicz-etal-2023-tokenization}.
See Appendix~\ref{sec:encoder_architecture} for brief model descriptions.
This allowed us to run correlation analyses without retraining the models, instead testing our metrics against an existing set of experiments.
The downstream results were obtained by fine-tuning the models on a given source language
(any of the available languages for the task) and evaluating on a target language,
resulting in many data points.
The tasks tested are XNLI \citep{conneau-etal-2018-xnli},
part-of-speech tagging (POS) and
dependency tagging (UD) (both based on~\citealp{zeman-etal-2019-ud25}), 
and named entity recognition (NER; \citealp{pan-etal-2017-cross}).
We always use Spearman's rank correlation to estimate the metrics' predictive power, following the previous work.

\subsection{Cross-lingual embedding alignment}\label{subsec:def-cla}

We measure
cross-lingual alignment between a language pair as retrieval accuracy on the Tatoeba dataset \citep{artetxe-schwenk-2019-massively}
as well as the FLORES-200 development set.
Following \citet{jones-etal-2021-massively}, we additionally compute average margin distances on the latter, that is, how much closer the correct match is to the source sentence than other target-side sentences are.
We do not compute word-level embedding alignment scores.

For encoder models, we create sentence embeddings by feeding the sentence to the model and averaging the encoder representations from layer 7 (with attention mask applied).
The reasoning is that the middle layers in XLM-R and similar encoder models, such as the ones we use, have been found to be more cross-lingually aligned
than the output layers \citep[e.g.][]{muller-etal-2021-first}.
For decoder models, we follow \citet{jiang2023scalingsentenceembeddingslarge} in using the prompt
``This sentence: \{sentence\} means in one word:'', then taking the last token representation of the last hidden layer as the sentence embedding.

\section{Results and Discussion}

\subsection{Main results}\label{subsec:main-results}

\begin{table}[t]

\footnotesize
\centering
\setlength{\tabcolsep}{4.5pt}
\begin{tabular}{l ccc@{\hskip 10pt}ccc@{\hskip 10pt}ccc}
\toprule
\multirow{2}{*}{Task} & \multicolumn{3}{c}{JSD} & \multicolumn{3}{c}{one-to-one} & \multicolumn{3}{c}{eflomal} \\
\cmidrule(lr){2-4} \cmidrule(lr){5-7} \cmidrule(lr){8-10}
& all & $=$ & $\neq$ & all & $=$ & $\neq$ & all & $=$ & $\neq$ \\
\midrule
XNLI  & \corr{0.32705537480624125}{-.33} & \corr{0.5670424483986245}{-.57} & \corr{0.3962737225709869}{\bf -.40} & \corr{0.29486428004102955}{\hphantom{-}.29} & \corr{0.5013683634373289}{\hphantom{-}.50} & \corr{0.21166660493168038}{\hphantom{-}.21} & \corr{0.4541827018801494}{\bf -.45} & \corr{0.6010430594244608}{\bf -.60} & \corr{0.3809108295618151}{-.38} \\
POS  & \corr{0.4470465041220072}{-.45} & \corr{0.6427212277787118}{\bf -.64} & \corr{0.4518107260414783}{-.45} & \corr{0.32066740982870195}{\hphantom{-}.32} & \corr{0.35960591133004927}{\hphantom{-}.36} & \corr{0.2900335205219909}{\hphantom{-}.29} & \corr{0.6406349884355937}{\bf -.64} & \corr{0.5012348141550704}{-.50} & \corr{0.6431149551523524}{\bf -.64} \\
UD  & \corr{0.2272662636409092}{-.23} & \corr{0.24787542231738713}{-.25} & \corr{0.24526885371739598}{-.25} & \corr{0.16003998342107087}{\hphantom{-}.16} & \corr{0.33278598795840175}{\hphantom{-}.33} & \corr{0.12654130128388702}{\hphantom{-}.13} & \corr{0.40921728295269644}{\bf -.41} & \corr{0.36084519443548835}{\bf -.36} & \corr{0.41551787295698384}{\bf -.42} \\
NER  & \corr{0.6288810824082914}{\bf -.63} & \corr{0.24690842283616005}{-.25} & \corr{0.4901424878565244}{\bf -.49} & \corr{0.28648233259060296}{\hphantom{-}.29} & \corr{0.3486095661846496}{\bf \hphantom{-}.35} & \corr{0.25094399137412293}{\hphantom{-}.25} & \corr{0.5164127888382387}{-.52} & \corr{0.2103624107918187}{-.21} & \corr{0.48112651723826266}{-.48} \\

\bottomrule
\end{tabular}

(a) Unigram

\vspace{5pt}

\begin{tabular}{l ccc@{\hskip 10pt}ccc@{\hskip 10pt}ccc}
\toprule
\multirow{2}{*}{Task} & \multicolumn{3}{c}{JSD} & \multicolumn{3}{c}{one-to-one} & \multicolumn{3}{c}{eflomal} \\
\cmidrule(lr){2-4} \cmidrule(lr){5-7} \cmidrule(lr){8-10}
& all & $=$ & $\neq$ & all & $=$ & $\neq$ & all & $=$ & $\neq$ \\
\midrule
XNLI  & \corr{0.5500301354267327}{\bf -.55} & \corr{0.4463951189521087}{-.45} & \corr{0.40408864401690736}{\bf -.40} & \corr{0.11453009246856921}{\hphantom{-}.11} & \corr{0.4625068418171866}{\bf \hphantom{-}.46} & \corr{0.05311941788014686}{\hphantom{-}.05} & \corr{0.44160198880160495}{-.44} & \corr{0.39374901155726544}{-.39} & \corr{0.29397695850178673}{-.29} \\
POS  & \corr{0.17227924150740712}{-.17} & \corr{0.6503987851071265}{\bf -.65} & \corr{0.07991243237853478}{-.08} & \corr{0.3528880268918219}{\hphantom{-}.35} & \corr{0.4395183360700602}{\hphantom{-}.44} & \corr{0.3329702643456619}{\hphantom{-}.33} & \corr{0.48753891439749886}{\bf -.49} & \corr{0.5231706922362551}{-.52} & \corr{0.46249046399642263}{\bf -.46} \\
UD  & \corr{0.15656120384151526}{-.16} & \corr{0.3005215297122304}{-.30} & \corr{0.14598684532143233}{-.15} & \corr{0.1847619225228333}{\hphantom{-}.18} & \corr{0.28735632183908044}{\hphantom{-}.29} & \corr{0.19114517333755224}{\hphantom{-}.19} & \corr{0.32838166067096686}{\bf -.33} & \corr{0.3575548127233106}{\bf -.36} & \corr{0.3201852377420793}{\bf -.32} \\
NER  & \corr{0.5131946615211677}{-.51} & \corr{0.37615651421248936}{-.38} & \corr{0.2984246755675126}{-.30} & \corr{0.29932113501342816}{\hphantom{-}.30} & \corr{0.5261401557285873}{\bf \hphantom{-}.53} & \corr{0.2828958974597811}{\hphantom{-}.28} & \corr{0.5738643304671307}{\bf -.57} & \corr{0.2549307181629667}{-.25} & \corr{0.5230900173043488}{\bf -.52} \\

\bottomrule
\end{tabular}

(b) BPE

\vspace{5pt}

\begin{tabular}{l ccc@{\hskip 10pt}ccc@{\hskip 10pt}ccc}
\toprule
\multirow{2}{*}{Task} & \multicolumn{3}{c}{JSD} & \multicolumn{3}{c}{one-to-one} & \multicolumn{3}{c}{eflomal} \\
\cmidrule(lr){2-4} \cmidrule(lr){5-7} \cmidrule(lr){8-10}
& all & $=$ & $\neq$ & all & $=$ & $\neq$ & all & $=$ & $\neq$ \\
\midrule
XNLI  & \corr{0.4507226442455602}{\bf -.45} & \corr{0.438717561623694}{\bf -.44} & \corr{0.4276799004135657}{\bf -.43} & \corr{0.06999578184951909}{-.07} & \corr{0.3448275862068966}{\hphantom{-}.34} & \corr{0.2262795053264527}{-.23} & \corr{0.35846844619079316}{-.36} & \corr{0.4343303860074571}{-.43} & \corr{0.22372797932156918}{-.22} \\
POS  & \corr{0.2057630630846998}{-.21} & \corr{0.6898833656532588}{\bf -.69} & \corr{0.11023558578565013}{-.11} & \corr{0.10519607066096628}{\hphantom{-}.11} & \corr{0.23371647509578544}{\hphantom{-}.23} & \corr{0.05830147605798354}{\hphantom{-}.06} & \corr{0.5363006387463846}{\bf -.54} & \corr{0.5100091653875443}{-.51} & \corr{0.5109600439457852}{\bf -.51} \\
UD  & \corr{0.17835161106603833}{-.18} & \corr{0.17439023074541837}{-.17} & \corr{0.16273364808610852}{-.16} & \corr{0.008905818013977407}{\hphantom{-}.01} & \corr{0.039956212370005476}{\hphantom{-}.04} & \corr{0.0026576041402034505}{-.00} & \corr{0.38386476704973804}{\bf -.38} & \corr{0.3301349651218297}{\bf -.33} & \corr{0.38619879747725644}{\bf -.39} \\
NER  & \corr{0.38373709728881145}{-.38} & \corr{0.32445727766195764}{\bf -.32} & \corr{0.08591988530953415}{-.09} & \corr{0.11305713045601282}{\hphantom{-}.11} & \corr{0.23203559510567295}{\hphantom{-}.23} & \corr{0.07801022071494197}{\hphantom{-}.08} & \corr{0.48215278853858723}{\bf -.48} & \corr{0.2745407734062718}{-.27} & \corr{0.4224916576006504}{\bf -.42} \\

\bottomrule
\end{tabular}

(c) TokMix

\caption{Spearman's rank correlation of downstream transfer with JSD, proportion of one-to-one alignment, and eflomal score, for language pairs with the same ($=$) and with a different script ($\neq$).
}
\label{tab:tokeniser-vs-downstream}
\end{table}

Table~\ref{tab:tokeniser-vs-downstream} shows that eflomal score is better than JSD at predicting downstream transfer performance in the multilingual encoder models from \citet{limisiewicz-etal-2023-tokenization}.
This holds across all three tokenisation types, particularly for the word-level tasks.
XNLI seems to behave differently, possibly because it is a sentence-level task in contrast with the other three, or because it has results available for fewer, mostly higher-resource, language pairs.
Note also that XNLI transfer results were quite low in absolute terms.

Intuitively, JSD clusters language pairs with different scripts very closely together, even when they have markedly different transfer performance (see visualisations in App. Fig.~\ref{fig:unigram-eflomal-over-jsd}--\ref{fig:tokmix-eflomal-over-jsd}).
Eflomal score is not confounded by the different scripts, yielding better rankings within that group, and usually a
better overall ranking.
Meanwhile, the proportion of one-to-one alignments shows weaker or no correlation.
This implies that the proportion of one-to-one alignments may be too simplistic here, while the eflomal score, as an estimate of log-probability, captures more nuance.

Table~\ref{tab:tokeniser-vs-embedding-alignment} lists correlations of JSD and eflomal score with three measures of embedding similarity (retrieval on Tatoeba and FLORES-200, and average margin on FLORES-200).
These results are for the BPE model.
The underlying distributions are shown in Fig.~\ref{fig:bpe-metrics-vs-cla}.
We see that JSD gives clear correlations for all three measures in \textit{same-script} language pairs, while eflomal score correlates more strongly on \textit{different-script} language pairs.

All the correlations are much stronger on the FLORES dataset, likely because this dataset was used to calculate the tokeniser metrics in the first place.
We can therefore see these as a kind of upper bound on how well the tokeniser metrics can predict cross-lingual alignment.
The fact that the eflomal score is less predictive in the same-script group may indicate that the model does rely on more literal token matching when that information is available.
To the extent that the behaviour differs from what is seen in Table~\ref{tab:tokeniser-vs-downstream}, this underscores that cross-lingual embedding alignment, as measured by similarity, is just one factor in the cross-lingual transfer ability of the model.

\begin{table}[t]

\footnotesize\centering
\setlength{\tabcolsep}{4.5pt}
\begin{tabular}{l ccc@{\hskip 10pt}ccc}
\toprule
\multirow{2}{*}{Task} & \multicolumn{3}{c}{JSD} & \multicolumn{3}{c}{eflomal} \\
\cmidrule(lr{10pt}){2-4} \cmidrule(lr){5-7}
& all & $=$ & $\neq$ & all & $=$ & $\neq$ \\
\midrule
F1 Flores   & \corr{0.7917754931925537}{-.79} & \corr{0.6969696969696969}{\bf -.70} & \corr{0.6678991596638655}{-.67} & \corr{0.825062517365935}{\bf -.83} & \corr{0.6242424242424242}{-.62} & \corr{0.8134933973589434}{\bf -.81} \\
Avg mgn Flores  & \corr{0.736093359266463}{-.74} & \corr{0.7212121212121211}{\bf -.72} & \corr{0.5878991596638654}{-.59} & \corr{0.8026674076132262}{\bf -.80} & \corr{0.4545454545454545}{-.45} & \corr{0.7933253301320528}{\bf -.79} \\
Tatoeba  & \corr{0.33378096461282464}{\bf -.33} & \corr{0.4588773602950718}{\bf -.46} & \corr{0.18936338104125688}{-.19} & \corr{0.32558510439095356}{-.33} & \corr{0.2660417380699054}{-.27} & \corr{0.24416498319842914}{\bf -.24} \\

\bottomrule
\end{tabular}

\caption{Spearman's rank correlation of embedding alignment with JSD and eflomal scores, on the BPE tokenizer/model.
We show overall correlations (all), same-script ($=$), and different script ($\neq$) pairs.}
\label{tab:tokeniser-vs-embedding-alignment}
\end{table}

\subsection{Is data size a confounder?}\label{subsec:data-size}

Table~\ref{tab:datasize-vs-downstream} shows data size in the trained encoders (and tokenizers), correlated with downstream transfer performance from English.
Here, we consider only the pairs where English is the source language because English is generally the most dominant language, and there is some research suggesting that models ``work'' in English \citep{wendler-etal-2024-llamas}.
This correlates very well for XNLI, but much less in the other tasks.
Again, XNLI stands out as a sentence-level task with fewer overall language pairs and relatively low transfer performance, so this result should be taken with a grain of salt.
Overall, the correlations suggest that there is indeed a connection between data size and transfer ability, but data size cannot account for the whole effect.
See also Table~\ref{tab:datasize-vs-metrics} in the Appendix.

\begin{table}[t]

\centering\footnotesize
\begin{tabularx}{170pt}{l *4{>{\Centering}X}}
\toprule
Model & XNLI & POS & UD & NER \\ \midrule
Unigram & \C{.87} & \C{.37} & \C{.33} & \C{.34} \\
BPE & \C{.80} & \C{.37} & \C{.49} & \C{.33} \\
TokMix & \C{.81} & \C{.34} & \C{.54} & \C{.26} \\
\bottomrule
\end{tabularx}

\caption{
Rank correlation of downstream transfer from English with training size of the target language.}
\label{tab:datasize-vs-downstream}
\end{table}

\subsection{What about decoders?}

We additionally experiment with Mistral-7B-v0.1, Aya23-8B, and Llama-3-8B-Instruct, varying the model type, as well as the amount of multilinguality in pre- and post-training.
For these, we calculate alignability scores, JSD, and representation alignment for a subset of language pairs.
Table~\ref{tab:aya-cla-vs-metrics} shows rank correlation results.
In Mistral, eflomal is still more predictive of overall representation alignment than JSD, while in Aya23 and Llama3, the opposite is true.
This may suggest that cross-linguality in these decoder models works differently than in encoder models, or that they \textit{do} rely more on literal token matches for their cross-linguality.
Nevertheless, in Llama3-8B-Instruct, the eflomal score shows an unusually high correlation for same-script language pairs.
Note also that absolute retrieval performance from the Mistral and Llama3 representations is quite low---Aya23 performs better.
The corresponding visualisations are shown in Appendix~\ref{subsec:app-decoders}.

\begin{table}[t]

\footnotesize\centering
\setlength{\tabcolsep}{4.5pt}

\begin{tabular}{ll ccc@{\hskip 10pt}ccc}
\toprule
\multirow{2}{*}{Model} & \multirow{2}{*}{Task} & \multicolumn{3}{c}{JSD} & \multicolumn{3}{c}{eflomal} \\

\cmidrule(lr{10pt}){3-5} \cmidrule(lr){6-8}
& & all & $=$ & $\neq$ & all & $=$ & $\neq$ \\
\midrule

\multirow{2}{*}{Aya23}
    & F1  & \corr{0.6805916305916307}{\bf -.68} & \corr{0.30952380952380953}{\bf \hphantom{-}.31} & \corr{0.7309204440333026}{\bf -.73} & \corr{0.48766233766233774}{-.49} & \corr{0.261904761904762}{-.26} & \corr{0.4299259944495837}{-.43} \\
    & Avg mgn  & \corr{0.6507215007215009}{\bf -.65} & \corr{0.30952380952380953}{\bf \hphantom{-}.31} & \corr{0.6722941720629046}{\bf -.67} & \corr{0.4286435786435787}{-.43} & \corr{0.261904761904762}{-.26} & \corr{0.3648242368177613}{-.36} \\

\midrule
\multirow{2}{*}{LLaMA3}
    & F1    & \corr{0.5872294372294373}{\bf -.59} & \corr{0.261904761904762}{-.26} & \corr{0.4498149861239593}{\bf -.45} & \corr{0.3201298701298702}{-.32} & \corr{0.5000000000000001}{\bf -.50} & \corr{0.17772895467160038}{-.18} \\
    & Avg mgn   & \corr{0.32958152958152964}{\bf -.33} & \corr{0.7380952380952381}{-.74} & \corr{0.016304347826086956}{\bf -.02} & \corr{0.20952380952380956}{-.21} & \corr{0.880952380952381}{\bf -.88} & \corr{0.016188714153561518}{-.02} \\

\midrule

\multirow{2}{*}{Mistral}
    & F1    & \corr{0.20110875956544533}{-.20} & \corr{0.04761904761904763}{-.05} & \corr{0.1627960610117019}{\hphantom{-}.16} & \corr{0.587077916606345}{\bf -.59} & \corr{0.6666666666666669}{\bf -.67} & \corr{0.5468261053812732}{\bf -.55} \\
    & Avg mgn   & \corr{0.21580086580086585}{-.22} & \corr{0.2380952380952381}{\bf \hphantom{-}.24} & \corr{0.1335568917668825}{\hphantom{-}.13} & \corr{0.7396825396825398}{\bf -.74} & \corr{0.2380952380952381}{\bf -.24} & \corr{0.7630666049953747}{\bf -.76} \\

\bottomrule
\end{tabular}

\caption{Spearman's rank correlation of embedding alignment with JSD and eflomal scores, on decoders.
We show overall correlations (all), same-script ($=$), and different script ($\neq$) pairs.
}
\label{tab:aya-cla-vs-metrics}
\end{table}

\section{Future Work}\label{sec:future}

We showed here that good tokeniser alignability correlates well with crosslinguality, an important factor for the performance of multilingual language models.
Hence, the eflomal score may be applied to improve vocabulary learning for fairer multilingual tokenisers (see also \citealp{ahia2024magnetimprovingmultilingualfairness,limisiewicz-etal-2024-myte}).
However, a naive implementation, where alignability score is checked at every decision point (merges for BPE, or pruning tokens for Unigram), is far too intensive.
Therefore, future work in this area will require finding suitable approximations, like calculating alignability score difference for some fraction (e.g., on the order of 10\%) of all candidate tokens at a time.

\section{Conclusion}

We have proposed a new metric for describing the quality of a multilingual tokenisation,
with implications for cross-lingual alignment in multilingual pre-trained models: token alignability.
This metric is particularly relevant for language pairs with different scripts and thus no literal token overlap.
We showed correlations with transfer performance on
downstream classification tasks, as well as with measures of cross-lingual alignment.
These findings show the potential of our token alignability metric to guide the development of robust multilingual tokenisers and to identify suitable language pairs for cross-lingual transfer.

\section*{Limitations}

Our study has focused on a relatively small set of models.
We do not have extensive cross-lingual transfer experiments for decoder models because fine-tuning each model on any number of languages would take too much compute.
Some of the downstream results from the previous work (particularly for XNLI) were quite poor in absolute terms, so they may not entirely reflect the situation in a higher-performance model.
While alignability score for one language pair is not very time-consuming to compute (and can be done on CPU), the time adds up quickly for a broader set of language pairs.
In its present formulation, alignability is also a corpus-wide score,
meaning it would require reformulating for word-level tasks.

\section*{Acknowledgments}

Thank you to Jindra Helcl for helpful discussions about this research.
KH is supported by the Munich Center for Machine Learning, and did much of the work on this project during a research visit to Prague.
The work at CUNI was supported by the Charles University project PRIMUS/23/SCI/023.

\bibliography{anthology,custom}

\begin{thebibliography}{44}
\providecommand{\natexlab}[1]{#1}

\bibitem[{Ahia et~al.(2024)Ahia, Kumar, Gonen, Hoffman, Limisiewicz, Tsvetkov, and Smith}]{ahia2024magnetimprovingmultilingualfairness}
Orevaoghene Ahia, Sachin Kumar, Hila Gonen, Valentin Hoffman, Tomasz Limisiewicz, Yulia Tsvetkov, and Noah~A. Smith. 2024.
\newblock \href {https://arxiv.org/abs/2407.08818} {Magnet: Improving the multilingual fairness of language models with adaptive gradient-based tokenization}.
\newblock \emph{preprint}, arXiv:2407.08818 [cs.CL].

\bibitem[{Alkaoud and Syed(2020)}]{alkaoud-syed-2020-importance}
Mohamed Alkaoud and Mairaj Syed. 2020.
\newblock \href {https://aclanthology.org/2020.wanlp-1.11} {On the importance of tokenization in {A}rabic embedding models}.
\newblock In \emph{Proceedings of the Fifth Arabic Natural Language Processing Workshop}, pages 119--129, Barcelona, Spain (Online). Association for Computational Linguistics.

\bibitem[{Artetxe and Schwenk(2019)}]{artetxe-schwenk-2019-massively}
Mikel Artetxe and Holger Schwenk. 2019.
\newblock \href {https://doi.org/10.1162/tacl_a_00288} {Massively multilingual sentence embeddings for zero-shot cross-lingual transfer and beyond}.
\newblock \emph{Transactions of the Association for Computational Linguistics}, 7:597--610.

\bibitem[{Bahdanau et~al.(2015)Bahdanau, Cho, and Bengio}]{bahdanau2015attention}
Dzmitry Bahdanau, {Kyung Hyun} Cho, and Yoshua Bengio. 2015.
\newblock Neural machine translation by jointly learning to align and translate.
\newblock In \emph{3rd International Conference on Learning Representations, ICLR 2015}.

\bibitem[{Batsuren et~al.(2024)Batsuren, Vylomova, Dankers, Delgerbaatar, Uzan, Pinter, and Bella}]{batsuren2024evaluatingsubwordtokenizationalien}
Khuyagbaatar Batsuren, Ekaterina Vylomova, Verna Dankers, Tsetsuukhei Delgerbaatar, Omri Uzan, Yuval Pinter, and Gábor Bella. 2024.
\newblock \href {https://arxiv.org/abs/2404.13292} {Evaluating subword tokenization: Alien subword composition and oov generalization challenge}.
\newblock \emph{preprint}, arXiv:2404.13292 [cs.CL].

\bibitem[{Brown et~al.(1993)Brown, {Della Pietra}, {Della Pietra}, and Mercer}]{brown93:tmo}
Peter Brown, Stephen {Della Pietra}, Vincent {Della Pietra}, and Robert Mercer. 1993.
\newblock \href {http://acl.ldc.upenn.edu/J/J93/J93-2003.pdf} {The mathematics of statistical machine translation{:} parameter estimation}.
\newblock \emph{Computational Linguistics}, 19(2):263--311.

\bibitem[{Conneau et~al.(2020)Conneau, Khandelwal, Goyal, Chaudhary, Wenzek, Guzm{\'a}n, Grave, Ott, Zettlemoyer, and Stoyanov}]{conneau-etal-2020-unsupervised}
Alexis Conneau, Kartikay Khandelwal, Naman Goyal, Vishrav Chaudhary, Guillaume Wenzek, Francisco Guzm{\'a}n, Edouard Grave, Myle Ott, Luke Zettlemoyer, and Veselin Stoyanov. 2020.
\newblock \href {https://doi.org/10.18653/v1/2020.acl-main.747} {Unsupervised cross-lingual representation learning at scale}.
\newblock In \emph{Proceedings of the 58th Annual Meeting of the Association for Computational Linguistics}, pages 8440--8451, Online. Association for Computational Linguistics.

\bibitem[{Conneau and Lample(2019)}]{conneau-lample-2019-cross}
Alexis Conneau and Guillaume Lample. 2019.
\newblock \href {https://proceedings.neurips.cc/paper_files/paper/2019/file/c04c19c2c2474dbf5f7ac4372c5b9af1-Paper.pdf} {Cross-lingual language model pretraining}.
\newblock In \emph{Advances in Neural Information Processing Systems}, volume~32. Curran Associates, Inc.

\bibitem[{Conneau et~al.(2018)Conneau, Rinott, Lample, Williams, Bowman, Schwenk, and Stoyanov}]{conneau-etal-2018-xnli}
Alexis Conneau, Ruty Rinott, Guillaume Lample, Adina Williams, Samuel Bowman, Holger Schwenk, and Veselin Stoyanov. 2018.
\newblock \href {https://doi.org/10.18653/v1/D18-1269} {{XNLI}: Evaluating cross-lingual sentence representations}.
\newblock In \emph{Proceedings of the 2018 Conference on Empirical Methods in Natural Language Processing}, pages 2475--2485, Brussels, Belgium. Association for Computational Linguistics.

\bibitem[{Deshpande et~al.(2022)Deshpande, Talukdar, and Narasimhan}]{deshpande-etal-2022-bert}
Ameet Deshpande, Partha Talukdar, and Karthik Narasimhan. 2022.
\newblock \href {https://doi.org/10.18653/v1/2022.naacl-main.264} {When is {BERT} multilingual? isolating crucial ingredients for cross-lingual transfer}.
\newblock In \emph{Proceedings of the 2022 Conference of the North American Chapter of the Association for Computational Linguistics: Human Language Technologies}, pages 3610--3623, Seattle, United States. Association for Computational Linguistics.

\bibitem[{Dufter and Sch{\"u}tze(2020)}]{dufter-schutze-2020-identifying}
Philipp Dufter and Hinrich Sch{\"u}tze. 2020.
\newblock \href {https://doi.org/10.18653/v1/2020.emnlp-main.358} {Identifying elements essential for {BERT}{'}s multilinguality}.
\newblock In \emph{Proceedings of the 2020 Conference on Empirical Methods in Natural Language Processing (EMNLP)}, pages 4423--4437, Online. Association for Computational Linguistics.

\bibitem[{El-Kishky et~al.(2020)El-Kishky, Chaudhary, Guzm{\'a}n, and Koehn}]{el-kishky-etal-2020-ccaligned}
Ahmed El-Kishky, Vishrav Chaudhary, Francisco Guzm{\'a}n, and Philipp Koehn. 2020.
\newblock \href {https://doi.org/10.18653/v1/2020.emnlp-main.480} {{CCA}ligned: A massive collection of cross-lingual web-document pairs}.
\newblock In \emph{Proceedings of the 2020 Conference on Empirical Methods in Natural Language Processing (EMNLP)}, pages 5960--5969, Online. Association for Computational Linguistics.

\bibitem[{Goyal et~al.(2022)Goyal, Gao, Chaudhary, Chen, Wenzek, Ju, Krishnan, Ranzato, Guzm{\'a}n, and Fan}]{goyal-etal-2022-flores}
Naman Goyal, Cynthia Gao, Vishrav Chaudhary, Peng-Jen Chen, Guillaume Wenzek, Da~Ju, Sanjana Krishnan, Marc{'}Aurelio Ranzato, Francisco Guzm{\'a}n, and Angela Fan. 2022.
\newblock \href {https://doi.org/10.1162/tacl_a_00474} {The {F}lores-101 evaluation benchmark for low-resource and multilingual machine translation}.
\newblock \emph{Transactions of the Association for Computational Linguistics}, 10:522--538.

\bibitem[{Guzm{\'a}n et~al.(2019)Guzm{\'a}n, Chen, Ott, Pino, Lample, Koehn, Chaudhary, and Ranzato}]{guzman-etal-2019-flores}
Francisco Guzm{\'a}n, Peng-Jen Chen, Myle Ott, Juan Pino, Guillaume Lample, Philipp Koehn, Vishrav Chaudhary, and Marc{'}Aurelio Ranzato. 2019.
\newblock \href {https://doi.org/10.18653/v1/D19-1632} {The {FLORES} evaluation datasets for low-resource machine translation: {N}epali{--}{E}nglish and {S}inhala{--}{E}nglish}.
\newblock In \emph{Proceedings of the 2019 Conference on Empirical Methods in Natural Language Processing and the 9th International Joint Conference on Natural Language Processing (EMNLP-IJCNLP)}, pages 6098--6111, Hong Kong, China. Association for Computational Linguistics.

\bibitem[{H{\"a}mmerl et~al.(2024)H{\"a}mmerl, Libovick{\'y}, and Fraser}]{hammerl-etal-2024-understanding}
Katharina H{\"a}mmerl, Jind{\v{r}}ich Libovick{\'y}, and Alexander Fraser. 2024.
\newblock \href {https://doi.org/10.18653/v1/2024.findings-acl.649} {Understanding cross-lingual {A}lignment{---}{A} survey}.
\newblock In \emph{Findings of the Association for Computational Linguistics ACL 2024}, pages 10922--10943, Bangkok, Thailand and virtual meeting. Association for Computational Linguistics.

\bibitem[{Hofmann et~al.(2022)Hofmann, Schuetze, and Pierrehumbert}]{hofmann-etal-2022-embarrassingly}
Valentin Hofmann, Hinrich Schuetze, and Janet Pierrehumbert. 2022.
\newblock \href {https://doi.org/10.18653/v1/2022.acl-short.43} {An embarrassingly simple method to mitigate undesirable properties of pretrained language model tokenizers}.
\newblock In \emph{Proceedings of the 60th Annual Meeting of the Association for Computational Linguistics (Volume 2: Short Papers)}, pages 385--393, Dublin, Ireland. Association for Computational Linguistics.

\bibitem[{Hu et~al.(2020)Hu, Ruder, Siddhant, Neubig, Firat, and Johnson}]{hu2020xtreme}
Junjie Hu, Sebastian Ruder, Aditya Siddhant, Graham Neubig, Orhan Firat, and Melvin Johnson. 2020.
\newblock \href {https://arxiv.org/abs/2003.11080} {Xtreme: A massively multilingual multi-task benchmark for evaluating cross-lingual generalization}.
\newblock \emph{preprint}, arXiv:2003.11080 [cs.CL].

\bibitem[{Hua et~al.(2024)Hua, Yun, and Pavlick}]{hua-etal-2024-mothello}
Tianze Hua, Tian Yun, and Ellie Pavlick. 2024.
\newblock \href {https://doi.org/10.18653/v1/2024.findings-naacl.103} {m{O}thello: When do cross-lingual representation alignment and cross-lingual transfer emerge in multilingual models?}
\newblock In \emph{Findings of the Association for Computational Linguistics: NAACL 2024}, pages 1585--1598, Mexico City, Mexico. Association for Computational Linguistics.

\bibitem[{Huang et~al.(2019)Huang, Liang, Duan, Gong, Shou, Jiang, and Zhou}]{huang-etal-2019-unicoder}
Haoyang Huang, Yaobo Liang, Nan Duan, Ming Gong, Linjun Shou, Daxin Jiang, and Ming Zhou. 2019.
\newblock \href {https://doi.org/10.18653/v1/D19-1252} {{U}nicoder: A universal language encoder by pre-training with multiple cross-lingual tasks}.
\newblock In \emph{Proceedings of the 2019 Conference on Empirical Methods in Natural Language Processing and the 9th International Joint Conference on Natural Language Processing (EMNLP-IJCNLP)}, pages 2485--2494, Hong Kong, China. Association for Computational Linguistics.

\bibitem[{Jiang et~al.(2023)Jiang, Huang, Luan, Wang, and Zhuang}]{jiang2023scalingsentenceembeddingslarge}
Ting Jiang, Shaohan Huang, Zhongzhi Luan, Deqing Wang, and Fuzhen Zhuang. 2023.
\newblock \href {https://arxiv.org/abs/2307.16645} {Scaling sentence embeddings with large language models}.
\newblock \emph{preprint}, arXiv:2307.16645 [cs.CL].

\bibitem[{Jones et~al.(2021)Jones, Wang, and Mahowald}]{jones-etal-2021-massively}
Alexander Jones, William~Yang Wang, and Kyle Mahowald. 2021.
\newblock \href {https://doi.org/10.18653/v1/2021.emnlp-main.471} {A massively multilingual analysis of cross-linguality in shared embedding space}.
\newblock In \emph{Proceedings of the 2021 Conference on Empirical Methods in Natural Language Processing}, pages 5833--5847, Online and Punta Cana, Dominican Republic. Association for Computational Linguistics.

\bibitem[{Kudo(2018)}]{kudo-2018-subword}
Taku Kudo. 2018.
\newblock \href {https://doi.org/10.18653/v1/P18-1007} {Subword regularization: Improving neural network translation models with multiple subword candidates}.
\newblock In \emph{Proceedings of the 56th Annual Meeting of the Association for Computational Linguistics (Volume 1: Long Papers)}, pages 66--75, Melbourne, Australia. Association for Computational Linguistics.

\bibitem[{Libovick{\'y} et~al.(2020)Libovick{\'y}, Rosa, and Fraser}]{libovicky-etal-2020-language}
Jind{\v{r}}ich Libovick{\'y}, Rudolf Rosa, and Alexander Fraser. 2020.
\newblock \href {https://doi.org/10.18653/v1/2020.findings-emnlp.150} {On the language neutrality of pre-trained multilingual representations}.
\newblock In \emph{Findings of the Association for Computational Linguistics: EMNLP 2020}, pages 1663--1674, Online. Association for Computational Linguistics.

\bibitem[{Limisiewicz et~al.(2023)Limisiewicz, Balhar, and Mare{\v{c}}ek}]{limisiewicz-etal-2023-tokenization}
Tomasz Limisiewicz, Ji{\v{r}}{\'\i} Balhar, and David Mare{\v{c}}ek. 2023.
\newblock \href {https://doi.org/10.18653/v1/2023.findings-acl.350} {Tokenization impacts multilingual language modeling: Assessing vocabulary allocation and overlap across languages}.
\newblock In \emph{Findings of the Association for Computational Linguistics: ACL 2023}, pages 5661--5681, Toronto, Canada. Association for Computational Linguistics.

\bibitem[{Limisiewicz et~al.(2024)Limisiewicz, Blevins, Gonen, Ahia, and Zettlemoyer}]{limisiewicz-etal-2024-myte}
Tomasz Limisiewicz, Terra Blevins, Hila Gonen, Orevaoghene Ahia, and Luke Zettlemoyer. 2024.
\newblock \href {https://doi.org/10.18653/v1/2024.acl-long.804} {{MYTE}: Morphology-driven byte encoding for better and fairer multilingual language modeling}.
\newblock In \emph{Proceedings of the 62nd Annual Meeting of the Association for Computational Linguistics (Volume 1: Long Papers)}, pages 15059--15076, Bangkok, Thailand. Association for Computational Linguistics.

\bibitem[{Lin(2006)}]{lin2006divergence}
J.~Lin. 2006.
\newblock \href {https://doi.org/10.1109/18.61115} {Divergence measures based on the shannon entropy}.
\newblock \emph{IEEE Trans. Inf. Theor.}, 37(1):145–151.

\bibitem[{Mofijul~Islam et~al.(2022)Mofijul~Islam, Aguilar, Ponnusamy, Solomon~Mathialagan, Ma, and Guo}]{mofijul-islam-etal-2022-vocabulary}
Md~Mofijul~Islam, Gustavo Aguilar, Pragaash Ponnusamy, Clint Solomon~Mathialagan, Chengyuan Ma, and Chenlei Guo. 2022.
\newblock \href {https://doi.org/10.18653/v1/2022.repl4nlp-1.10} {A vocabulary-free multilingual neural tokenizer for end-to-end task learning}.
\newblock In \emph{Proceedings of the 7th Workshop on Representation Learning for NLP}, pages 91--99, Dublin, Ireland. Association for Computational Linguistics.

\bibitem[{Muller et~al.(2021)Muller, Elazar, Sagot, and Seddah}]{muller-etal-2021-first}
Benjamin Muller, Yanai Elazar, Beno{\^\i}t Sagot, and Djam{\'e} Seddah. 2021.
\newblock \href {https://doi.org/10.18653/v1/2021.eacl-main.189} {First align, then predict: Understanding the cross-lingual ability of multilingual {BERT}}.
\newblock In \emph{Proceedings of the 16th Conference of the European Chapter of the Association for Computational Linguistics: Main Volume}, pages 2214--2231, Online. Association for Computational Linguistics.

\bibitem[{{\"O}stling and Tiedemann(2016)}]{oestling-tiedemann-2016-efficientWA}
Robert {\"O}stling and J{\"o}rg Tiedemann. 2016.
\newblock \href {https://doi.org/10.1515/pralin-2016-0013} {Efficient word alignment with markov chain monte carlo}.
\newblock \emph{The Prague Bulletin of Mathematical Linguistics}, 106:125 -- 146.

\bibitem[{Pan et~al.(2017)Pan, Zhang, May, Nothman, Knight, and Ji}]{pan-etal-2017-cross}
Xiaoman Pan, Boliang Zhang, Jonathan May, Joel Nothman, Kevin Knight, and Heng Ji. 2017.
\newblock \href {https://doi.org/10.18653/v1/P17-1178} {Cross-lingual name tagging and linking for 282 languages}.
\newblock In \emph{Proceedings of the 55th Annual Meeting of the Association for Computational Linguistics (Volume 1: Long Papers)}, pages 1946--1958, Vancouver, Canada. Association for Computational Linguistics.

\bibitem[{Pham et~al.(2024)Pham, Le, and Luu}]{pham-etal-2024-unibridge}
Trinh Pham, Khoi Le, and Anh~Tuan Luu. 2024.
\newblock \href {https://doi.org/10.18653/v1/2024.acl-long.174} {{U}ni{B}ridge: A unified approach to cross-lingual transfer learning for low-resource languages}.
\newblock In \emph{Proceedings of the 62nd Annual Meeting of the Association for Computational Linguistics (Volume 1: Long Papers)}, pages 3168--3184, Bangkok, Thailand. Association for Computational Linguistics.

\bibitem[{Schmidt et~al.(2024)Schmidt, Reddy, Zhang, Alameddine, Uzan, Pinter, and Tanner}]{schmidt2024tokenizationcompression}
Craig~W. Schmidt, Varshini Reddy, Haoran Zhang, Alec Alameddine, Omri Uzan, Yuval Pinter, and Chris Tanner. 2024.
\newblock \href {https://arxiv.org/abs/2402.18376} {Tokenization is more than compression}.
\newblock \emph{preprint}, arXiv:2402.18376 [cs.CL].

\bibitem[{Schuster et~al.(2019)Schuster, Gupta, Shah, and Lewis}]{schuster-etal-2019-cross-lingual}
Sebastian Schuster, Sonal Gupta, Rushin Shah, and Mike Lewis. 2019.
\newblock \href {https://doi.org/10.18653/v1/N19-1380} {Cross-lingual transfer learning for multilingual task oriented dialog}.
\newblock In \emph{Proceedings of the 2019 Conference of the North {A}merican Chapter of the Association for Computational Linguistics: Human Language Technologies, Volume 1 (Long and Short Papers)}, pages 3795--3805, Minneapolis, Minnesota. Association for Computational Linguistics.

\bibitem[{Schäfer et~al.(2024)Schäfer, Ravfogel, Hofmann, Pimentel, and Schlag}]{schäfer2024rolelanguageimbalancecrosslingual}
Anton Schäfer, Shauli Ravfogel, Thomas Hofmann, Tiago Pimentel, and Imanol Schlag. 2024.
\newblock \href {https://arxiv.org/abs/2404.07982} {The role of language imbalance in cross-lingual generalisation: Insights from cloned language experiments}.
\newblock \emph{preprint}, arXiv:2404.07982 [cs.CL].

\bibitem[{Sennrich et~al.(2016)Sennrich, Haddow, and Birch}]{sennrich-etal-2016-neural}
Rico Sennrich, Barry Haddow, and Alexandra Birch. 2016.
\newblock \href {https://doi.org/10.18653/v1/P16-1162} {Neural machine translation of rare words with subword units}.
\newblock In \emph{Proceedings of the 54th Annual Meeting of the Association for Computational Linguistics (Volume 1: Long Papers)}, pages 1715--1725, Berlin, Germany. Association for Computational Linguistics.

\bibitem[{Team et~al.(2022)Team, Costa-jussà, Cross, Çelebi, Elbayad, Heafield, Heffernan, Kalbassi, Lam, Licht, Maillard, Sun, Wang, Wenzek, Youngblood, Akula, Barrault, Gonzalez, Hansanti, Hoffman, Jarrett, Sadagopan, Rowe, Spruit, Tran, Andrews, Ayan, Bhosale, Edunov, Fan, Gao, Goswami, Guzmán, Koehn, Mourachko, Ropers, Saleem, Schwenk, and Wang}]{nllb2022}
NLLB Team, Marta~R. Costa-jussà, James Cross, Onur Çelebi, Maha Elbayad, Kenneth Heafield, Kevin Heffernan, Elahe Kalbassi, Janice Lam, Daniel Licht, Jean Maillard, Anna Sun, Skyler Wang, Guillaume Wenzek, Al~Youngblood, Bapi Akula, Loic Barrault, Gabriel~Mejia Gonzalez, Prangthip Hansanti, John Hoffman, Semarley Jarrett, Kaushik~Ram Sadagopan, Dirk Rowe, Shannon Spruit, Chau Tran, Pierre Andrews, Necip~Fazil Ayan, Shruti Bhosale, Sergey Edunov, Angela Fan, Cynthia Gao, Vedanuj Goswami, Francisco Guzmán, Philipp Koehn, Alexandre Mourachko, Christophe Ropers, Safiyyah Saleem, Holger Schwenk, and Jeff Wang. 2022.
\newblock \href {https://arxiv.org/abs/2207.04672} {No language left behind: Scaling human-centered machine translation}.
\newblock \emph{preprint}, arXiv:2207.04672 [cs.CL].

\bibitem[{Tiedemann(2012)}]{tiedemann-2012-parallel}
J{\"o}rg Tiedemann. 2012.
\newblock \href {http://www.lrec-conf.org/proceedings/lrec2012/pdf/463_Paper.pdf} {Parallel data, tools and interfaces in {OPUS}}.
\newblock In \emph{Proceedings of the Eighth International Conference on Language Resources and Evaluation ({LREC}'12)}, pages 2214--2218, Istanbul, Turkey. European Language Resources Association (ELRA).

\bibitem[{V{\'a}zquez et~al.(2019)V{\'a}zquez, Sulubacak, and Tiedemann}]{vazquez-etal-2019-university}
Ra{\'u}l V{\'a}zquez, Umut Sulubacak, and J{\"o}rg Tiedemann. 2019.
\newblock \href {https://doi.org/10.18653/v1/W19-5441} {The {U}niversity of {H}elsinki submission to the {WMT}19 parallel corpus filtering task}.
\newblock In \emph{Proceedings of the Fourth Conference on Machine Translation (Volume 3: Shared Task Papers, Day 2)}, pages 294--300, Florence, Italy. Association for Computational Linguistics.

\bibitem[{Wendler et~al.(2024)Wendler, Veselovsky, Monea, and West}]{wendler-etal-2024-llamas}
Chris Wendler, Veniamin Veselovsky, Giovanni Monea, and Robert West. 2024.
\newblock \href {https://doi.org/10.18653/v1/2024.acl-long.820} {Do llamas work in {E}nglish? on the latent language of multilingual transformers}.
\newblock In \emph{Proceedings of the 62nd Annual Meeting of the Association for Computational Linguistics (Volume 1: Long Papers)}, pages 15366--15394, Bangkok, Thailand. Association for Computational Linguistics.

\bibitem[{Wu and Dredze(2019)}]{wu-dredze-2019-beto}
Shijie Wu and Mark Dredze. 2019.
\newblock \href {https://doi.org/10.18653/v1/D19-1077} {Beto, bentz, becas: The surprising cross-lingual effectiveness of {BERT}}.
\newblock In \emph{Proceedings of the 2019 Conference on Empirical Methods in Natural Language Processing and the 9th International Joint Conference on Natural Language Processing (EMNLP-IJCNLP)}, pages 833--844, Hong Kong, China. Association for Computational Linguistics.

\bibitem[{Xue et~al.(2022)Xue, Barua, Constant, Al-Rfou, Narang, Kale, Roberts, and Raffel}]{xue-etal-2022-byt5}
Linting Xue, Aditya Barua, Noah Constant, Rami Al-Rfou, Sharan Narang, Mihir Kale, Adam Roberts, and Colin Raffel. 2022.
\newblock \href {https://doi.org/10.1162/tacl_a_00461} {{B}y{T}5: Towards a token-free future with pre-trained byte-to-byte models}.
\newblock \emph{Transactions of the Association for Computational Linguistics}, 10:291--306.

\bibitem[{Zeman et~al.(2019)Zeman, Nivre et~al.}]{zeman-etal-2019-ud25}
Daniel Zeman, Joakim Nivre, et~al. 2019.
\newblock \href {http://hdl.handle.net/11234/1-3105} {Universal dependencies 2.5}.
\newblock {LINDAT}/{CLARIAH}-{CZ} digital library at the Institute of Formal and Applied Linguistics ({{\'U}FAL}), Faculty of Mathematics and Physics, Charles University.

\bibitem[{Zhang et~al.(2020)Zhang, Williams, Titov, and Sennrich}]{zhang-etal-2020-improving}
Biao Zhang, Philip Williams, Ivan Titov, and Rico Sennrich. 2020.
\newblock \href {https://doi.org/10.18653/v1/2020.acl-main.148} {Improving massively multilingual neural machine translation and zero-shot translation}.
\newblock In \emph{Proceedings of the 58th Annual Meeting of the Association for Computational Linguistics}, pages 1628--1639, Online. Association for Computational Linguistics.

\bibitem[{Zouhar et~al.(2023)Zouhar, Meister, Gastaldi, Du, Sachan, and Cotterell}]{zouhar-etal-2023-tokenization}
Vil{\'e}m Zouhar, Clara Meister, Juan Gastaldi, Li~Du, Mrinmaya Sachan, and Ryan Cotterell. 2023.
\newblock \href {https://doi.org/10.18653/v1/2023.acl-long.284} {Tokenization and the noiseless channel}.
\newblock In \emph{Proceedings of the 61st Annual Meeting of the Association for Computational Linguistics (Volume 1: Long Papers)}, pages 5184--5207, Toronto, Canada. Association for Computational Linguistics.

\end{thebibliography}

\appendix

\section{Languages Included}
\label{sec:app-langs}

We start from a set of 20 languages, namely the ones used by \citet{limisiewicz-etal-2023-tokenization} for their tokenizers: Arabic~(ar), Turkish~(tr), Chinese~(zh), Greek~(el), Spanish~(es), English~(en), Swahili~(sw), Hindi~(hi), Marathi~(mr), Urdu~(ur), Tamil~(ta), Telugu~(te), Thai~(th), Russian~(ru), Bulgarian~(bg), Hebrew~(he), Georgian~(ka), Vietnamese~(vi), French~(fr), and German~(de).

This gives us up to 190 language pairs (before accounting for direction), but we typically do not calculate numbers for \textit{all} pairs, and each downstream task only has data available for some subset of the languages.
We do compute all language pairs with English as either the source or target language.
For non-English pairs, we compute token alignability for the product of these languages: ar, tr, zh, hi, ur, mr, ru, bg, vi, fr, es, ta, he.

\section{Encoder Details}
 \label{sec:encoder_architecture}

The encoders were trained by \citet{limisiewicz-etal-2023-tokenization}. The models' architecture is based on XLM-RoBERTa \cite{conneau-etal-2020-unsupervised}.
The size of the embeddings is 768, the number of attention layers is 8, and the number of attention heads is 6. 
The maximum sentence length is 128, and the vocabulary size in each tokenizer is 120000. 
The number of parameters is 150M, roughly half the size of XLM-R$_{base}$.
See \citet{limisiewicz-etal-2023-tokenization} for training details.
Their training corpus was a 10\% subset of CC-100, with a balancing factor of $\alpha=0.25$ (cf.~\citealp{conneau-lample-2019-cross}).
The model names BPE, Unigram, and TokMix are shorthand for their different vocabulary creation approaches.
For BPE and Unigram, they simply applied the respective algorithm to the training set of all 20 languages, until reaching the target vocabulary size of 120000.
For TokMix, they
trained Unigram LM tokenisers for each language separately, and merged them by averaging token probabilities across tokenisers, then sorting and trimming.
Our own experiments with these models were able to run on CPU.

\section{Additional Detail on Results}

\subsection{Graphs for Main Results}\label{subsec:extra-graphs-main}

\begin{figure*}[htb]
  \centering
  \includegraphics[width=\textwidth]{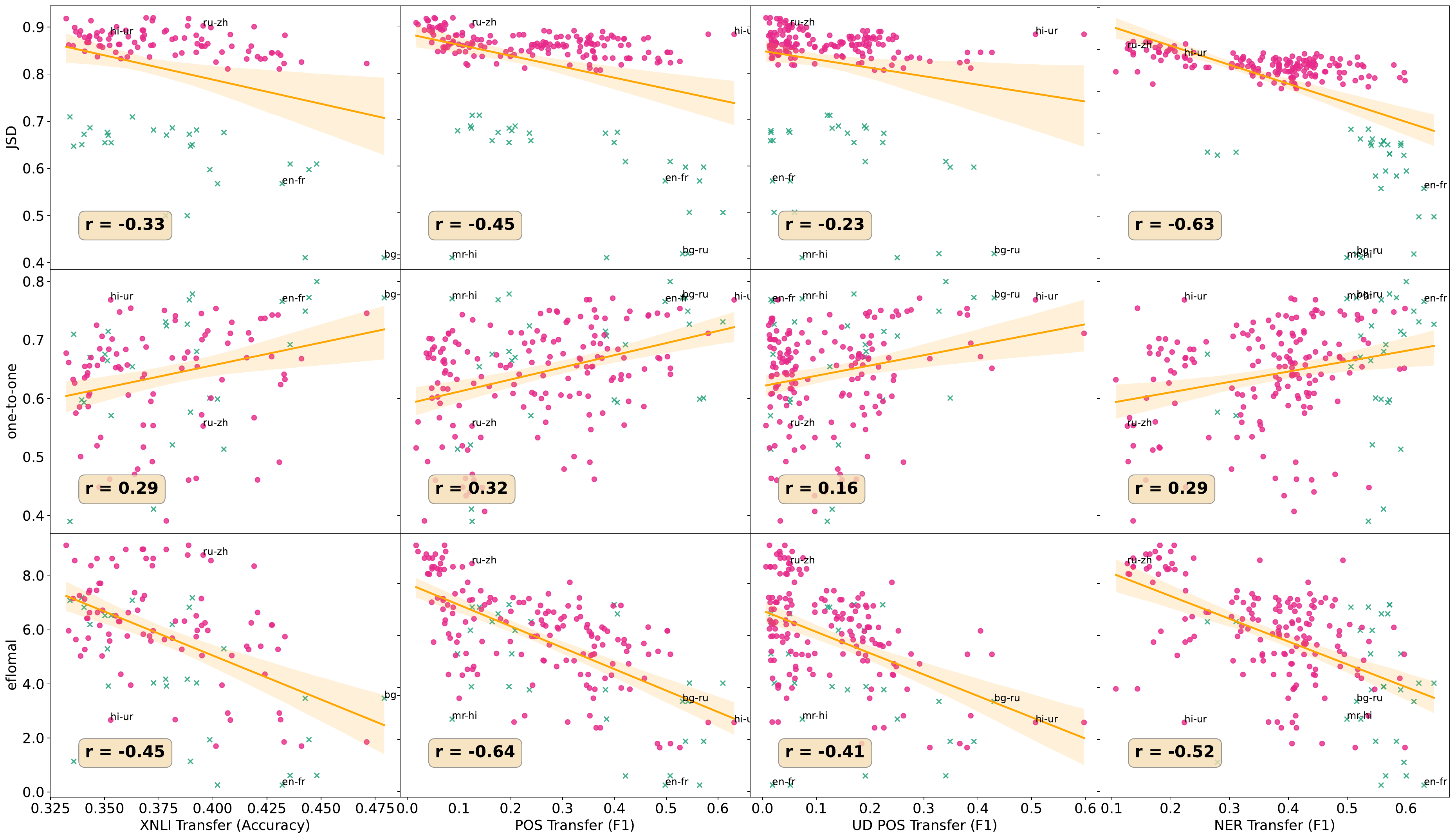}
  \caption{Unigram model: The eflomal score generally correlates better with downstream transfer than JSD. NER is the exception.
  Proportion of 1-1 token alignments, while it also breaks up the cluster of different-script language pairs, shows weaker or no correlations.
  }
  \label{fig:unigram-eflomal-over-jsd}
\end{figure*}

\begin{figure*}[htb]
  \centering
  \includegraphics[width=\textwidth]{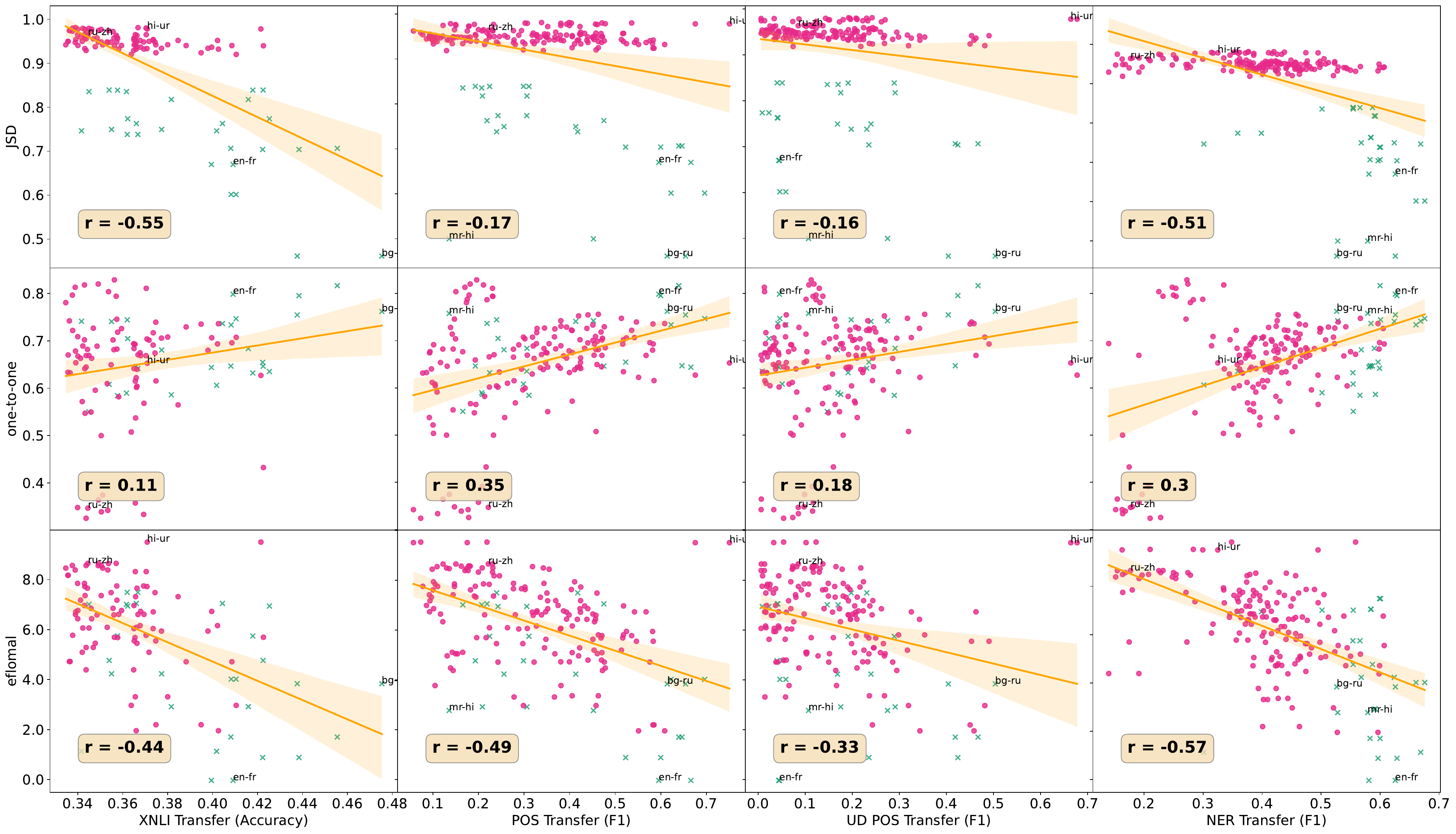}
  \caption{BPE model: The eflomal score correlates better with downstream transfer than JSD, with the exception of XNLI.
  Proportion of 1-1 token alignments, while it also breaks up the cluster of different-script language pairs, shows weaker or no correlations.
  }
  \label{fig:bpe-eflomal-over-jsd}
\end{figure*}

\begin{figure*}[htb]
  \centering
  \includegraphics[width=\textwidth]{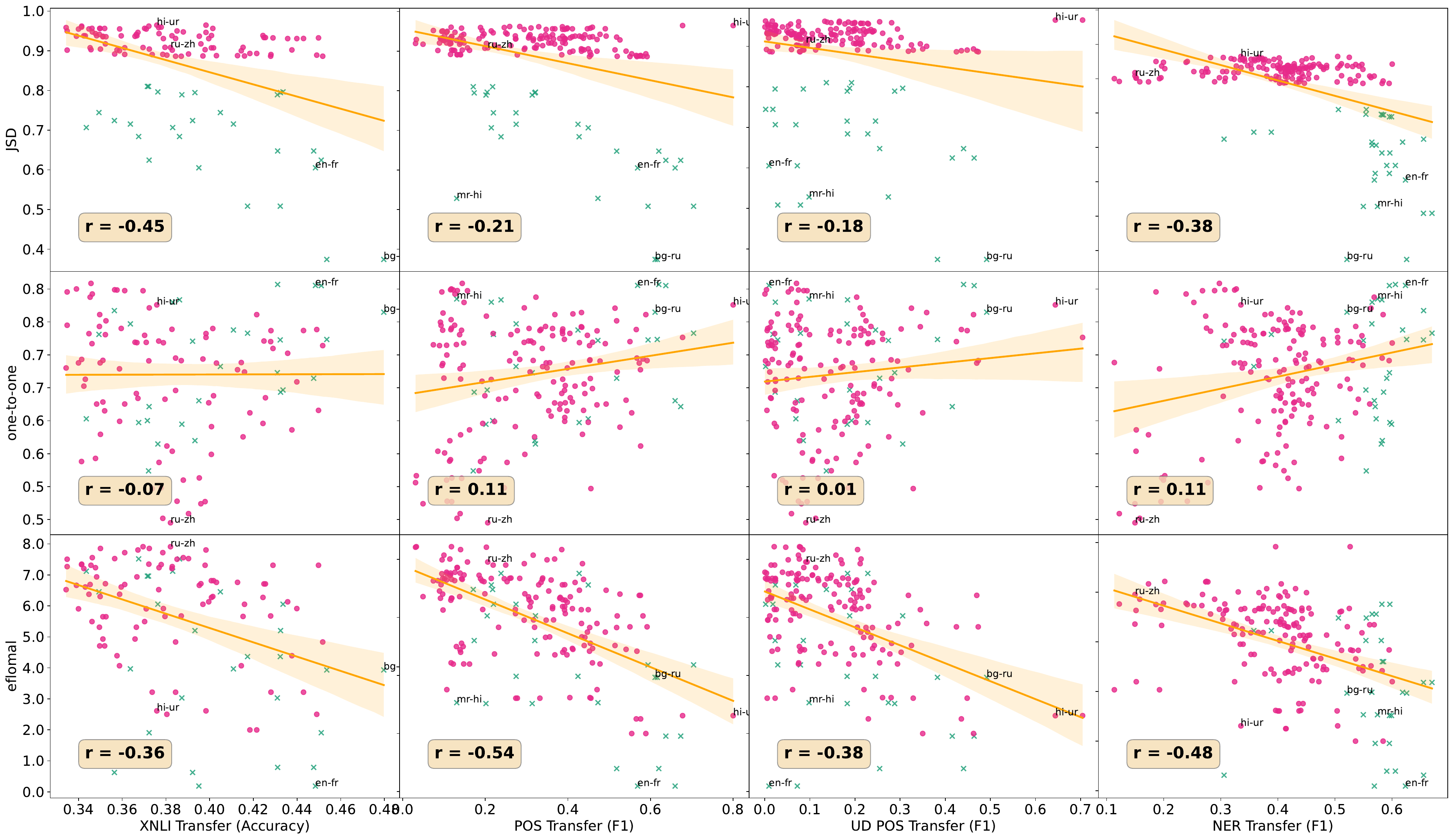}
  \caption{TokMix model: The eflomal score correlates better with downstream transfer than JSD, again with the exception of XNLI.
  Proportion of 1-1 token alignments, while it also breaks up the cluster of different-script language pairs, shows no correlations.
  }
  \label{fig:tokmix-eflomal-over-jsd}
\end{figure*}

\begin{figure*}
    \centering
    \includegraphics[width=\textwidth]{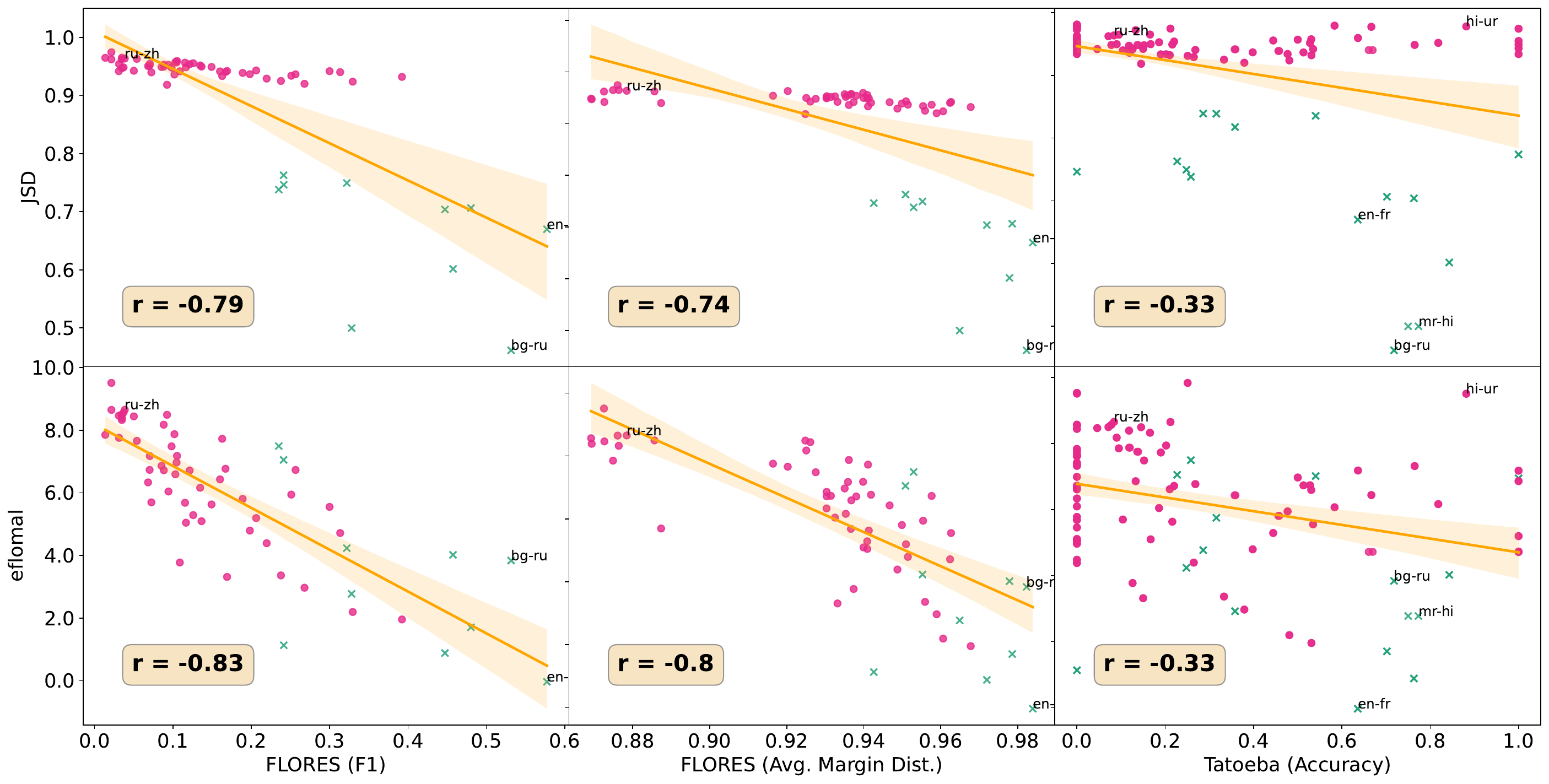}
\caption{BPE Model: Eflomal scores correlates well with cross-lingual embedding alignment. Nevertheless, both metrics perform similarly over the Tatoeba dataset.
}
\label{fig:bpe-metrics-vs-cla}
\end{figure*}

Figures~\ref{fig:unigram-eflomal-over-jsd},~\ref{fig:bpe-eflomal-over-jsd}, and~\ref{fig:tokmix-eflomal-over-jsd} visualise the distributions underlying Table~\ref{tab:tokeniser-vs-downstream}.
The sets of same- and different-script language pairs are colour-coded, and the overall correlations along with p-values are placed in the bottom left corner of each graph.
Similarly, Figure~\ref{fig:bpe-metrics-vs-cla} shows the distributions behind Table~\ref{tab:tokeniser-vs-embedding-alignment}.

\subsection{Analysis by Language Family}

Similarly to our analysis of scripts, we assign language \textit{pairs} to groups of same vs. different macro language families.
We do this because some language families have just one representative in our set, while Indo-European accounts for many of the languages.
We do not subdivide the macro language families for this analysis.

Table~\ref{tab:by-family} shows the correlations of eflomal score with downstream cross-lingual transfer, over different-script pairs.
We then split by same and different language families.
In several cases, we see very similar correlations as on different-script pairs in general.
XNLI stands out again, with pairs from the same language family tending to be more correlated across all tokenisers.

\begin{table}[t]

\footnotesize
\centering
\setlength{\tabcolsep}{4.5pt}
\begin{tabular}{l ccc@{\hskip 12pt}ccc@{\hskip 12pt}ccc}
\toprule
\multirow{2}{*}{Task} & \multicolumn{3}{c}{Unigram} & \multicolumn{3}{c}{BPE} & \multicolumn{3}{c}{TokMix} \\
\cmidrule(lr){2-4} \cmidrule(lr){5-7} \cmidrule(lr){8-10}
& all & $=$ & $\neq$ & all & $=$ & $\neq$ & all & $=$ & $\neq$ \\
\midrule
XNLI  & \corr{0.3809108295618151}{-.38} & \corr{0.5957999109809452}{-.60} & \corr{0.22493424068017584}{-.22} & \corr{0.29397695850178673}{-.29} & \corr{0.34480115955558166}{-.34} & \corr{0.26324925352280415}{-.26} & \corr{0.22372797932156918}{-.22} & \corr{0.4231688640995684}{-.42} & \corr{0.22577756613005404}{-.23} \\
POS  & \corr{0.6431149551523524}{-.64} & \corr{0.41763370895828206}{-.42} & \corr{0.6938120272539409}{-.69} & \corr{0.46249046399642263}{-.46} & \corr{0.23428232453757286}{-.23} & \corr{0.47773538200050264}{-.48} & \corr{0.5109600439457852}{-.51} & \corr{0.383231300351042}{-.38} & \corr{0.44364409367186686}{-.44} \\
UD  & \corr{0.41551787295698384}{-.42} & \corr{0.2990511049880749}{-.30} & \corr{0.40701834958710553}{-.41} & \corr{0.3201852377420793}{-.32} & \corr{0.08148950418698185}{-.08} & \corr{0.36936357835324063}{-.37} & \corr{0.38619879747725644}{-.39} & \corr{0.3317237835536101}{-.33} & \corr{0.33483400068332}{-.33} \\
NER  & \corr{0.48112651723826266}{-.48} & \corr{0.3201922499422448}{-.32} & \corr{0.5213832969100821}{-.52} & \corr{0.5230900173043488}{-.52} & \corr{0.5089250167149244}{-.51} & \corr{0.5133869874572256}{-.51} & \corr{0.4224916576006504}{-.42} & \corr{0.3267267856553518}{-.33} & \corr{0.3767321092436638}{-.38} \\

\bottomrule
\end{tabular}

\caption{Spearman's rank correlation of downstream transfer with JSD, proportion of one-to-one alignment, and eflomal score.
This analysis shows only language pairs that use \textit{different scripts}, further differentiated by whether they are in the same ($=$) or a different ($\neq$) \textit{language family}.
}
\label{tab:by-family}
\end{table}

\subsection{Data Size Correlated with Metrics}

Table~\ref{tab:datasize-vs-metrics} shows the correlations of target language pre-training data sizes with our tokeniser metrics.

\begin{table}[h]

\footnotesize\centering
\begin{tabularx}{.9\columnwidth}{l *3{>{\Centering}X}}
\toprule
& JSD & one-to-one & eflomal \\ \midrule
Unigram & \corr{0.3025516248337339}{-.30} & \corr{0.4890078587428955}{\hphantom{0}.49} & \corr{0.4362372265044535}{-.44} \\
BPE & \corr{0.4028158260867736}{-.40} & \corr{0.23834735561029616}{\hphantom{0}.24} & \corr{0.5356219172201858}{-.54} \\
TokMix & \corr{0.48373079551905124}{-.48} & \corr{0.29991309322181176}{\hphantom{0}.30} & \corr{0.5180317064740385}{-.52} \\
\bottomrule\end{tabularx}

\caption{Spearman's rank correlation of the target language pre-training data size with our metrics. Only pairs with English as the source language are considered for this table.}
\label{tab:datasize-vs-metrics}
\end{table}


\subsection{Graphs for Decoder Results}\label{subsec:app-decoders}

\begin{figure}
    \centering
    \includegraphics[width=\linewidth]{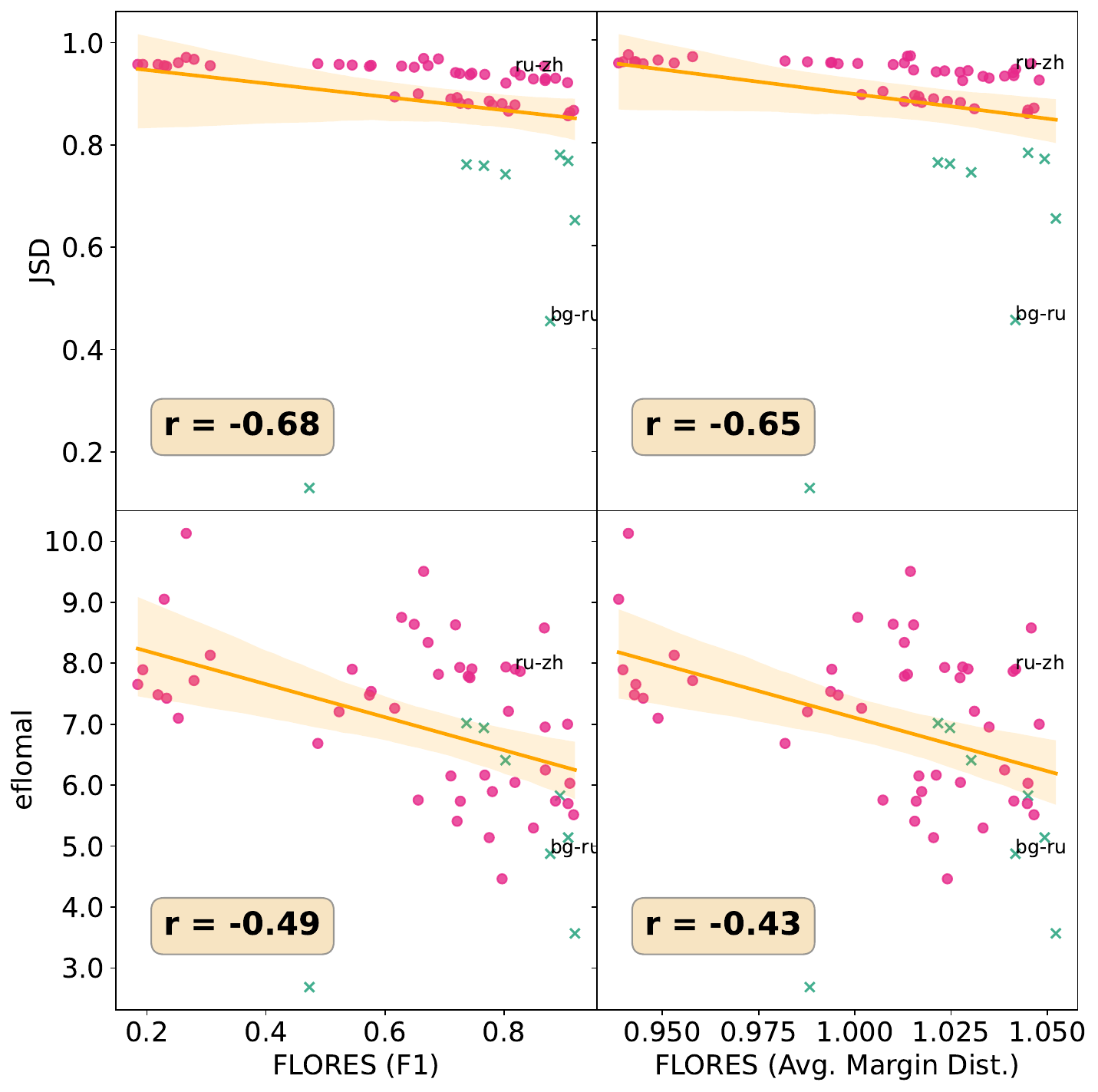}
\caption{Aya23: Spearman's rank correlation of cross-lingual embedding alignment with JSD and eflomal score.
}
\label{fig:metrics-vs-cla-aya}
\end{figure}

\begin{figure}
    \centering
    \includegraphics[width=\linewidth]{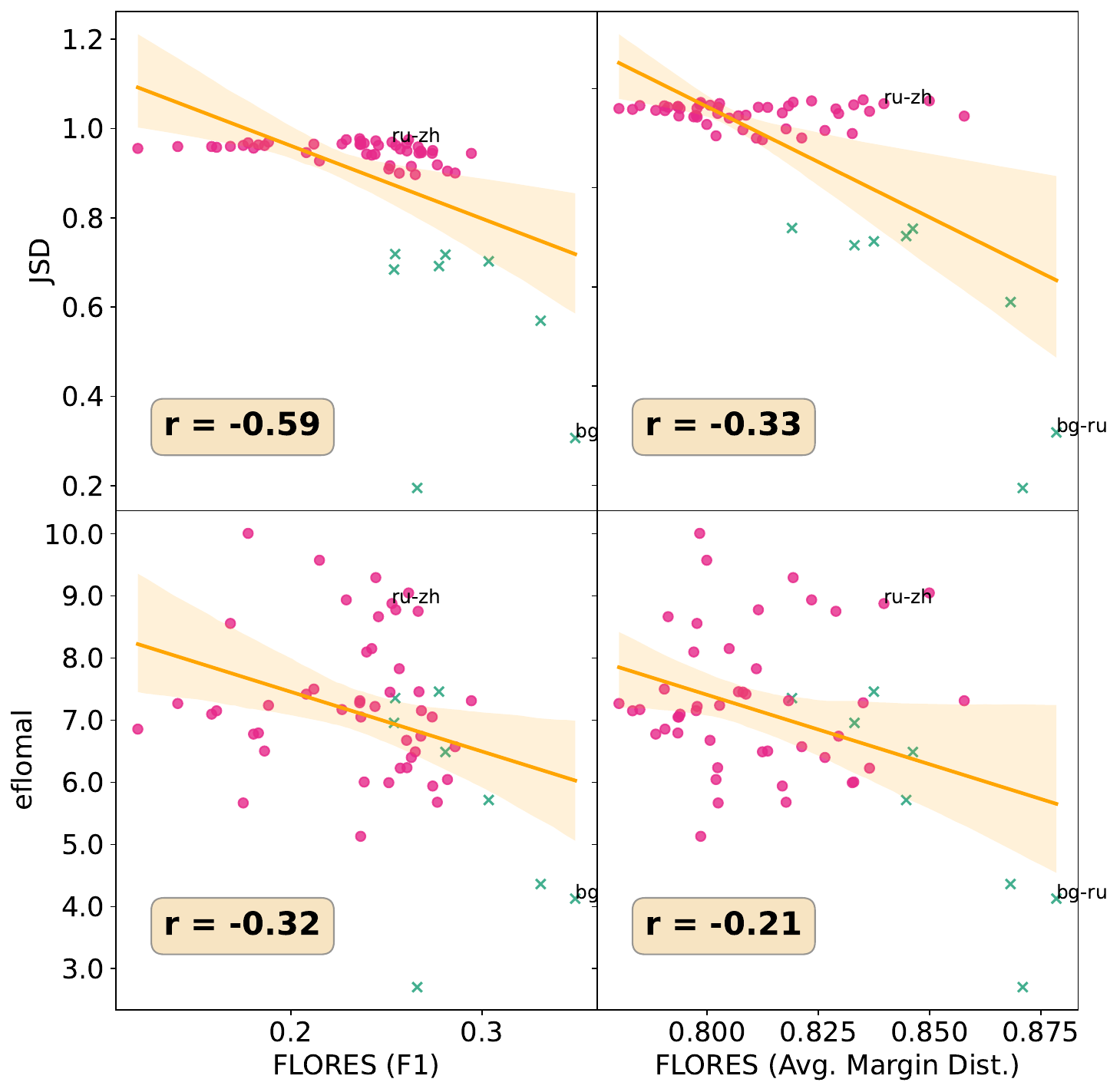}
\caption{Llama3: Spearman's rank correlation of cross-lingual embedding alignment with JSD and eflomal score.
}
\label{fig:metrics-vs-cla-llama}
\end{figure}

\begin{figure}
    \centering
    \includegraphics[width=\linewidth]{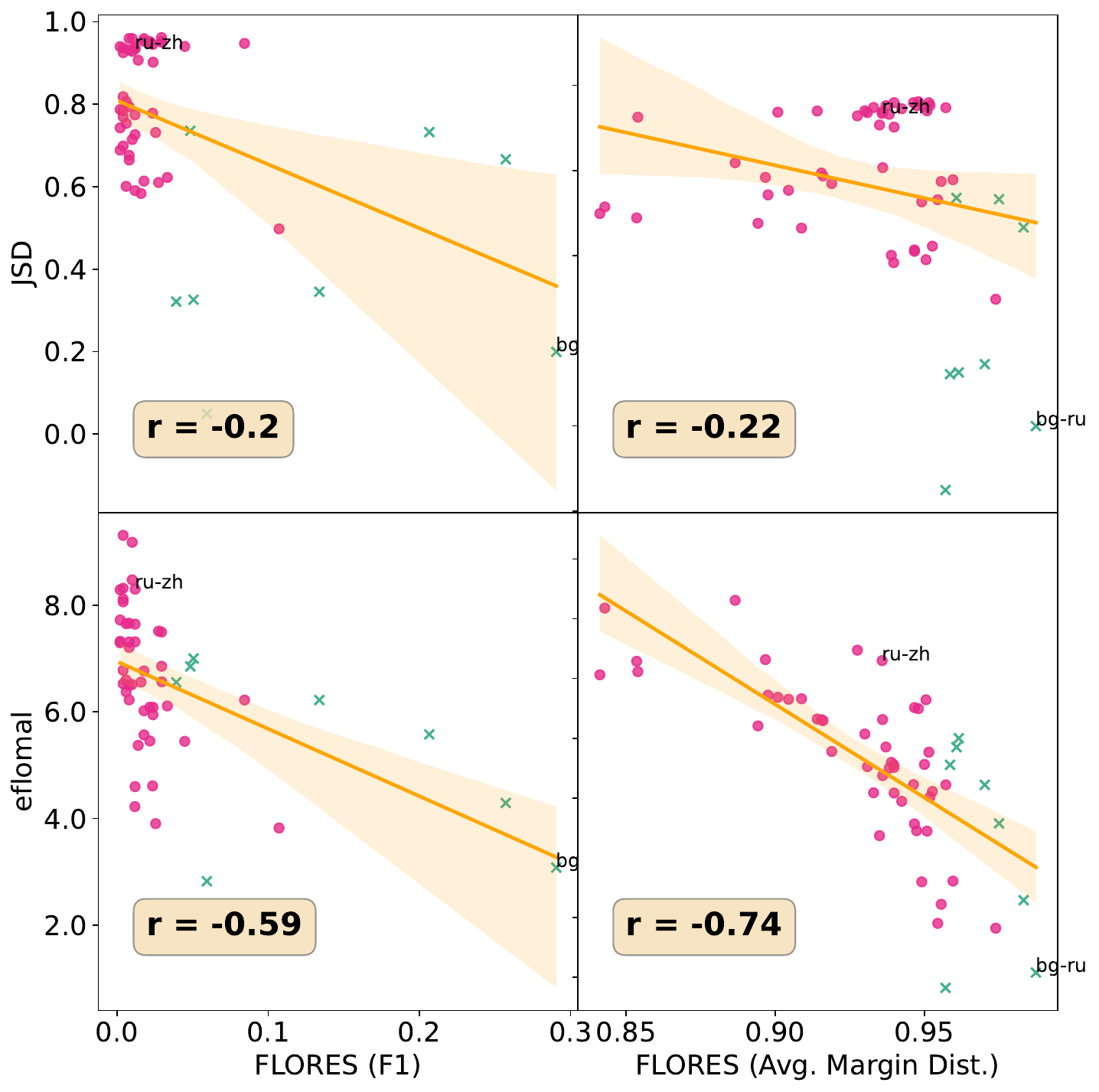}
\caption{Mistral: Spearman's rank correlation of cross-lingual embedding alignment with JSD and eflomal score.
}
\label{fig:metrics-vs-cla-mistral}
\end{figure}

The underlying distributions of Table~\ref{tab:aya-cla-vs-metrics} are visualised in Figure~\ref{fig:metrics-vs-cla-aya} for Aya23-8B,
Figure~\ref{fig:metrics-vs-cla-llama} for Llama-3-8B-Instruct, and Figure~\ref{fig:metrics-vs-cla-mistral} for Mistral.
Both in Llama3-8B-Instruct and Aya23-8B, JSD correlates more strongly with cross-lingual alignment of representations, but all correlations here are weaker than is the case in the encoder models.
For Mistral, eflomal score correlates more with cross-lingual alignment, which is in contrast to the other two decoder models.

Also, note that Aya23 shows decent retrieval performance, while the representations from Llama3 and Mistral both perform poorly on retrieval F1.

\end{document}